\documentclass[lettersize,journal]{IEEEtran}
\usepackage{amsmath,amsfonts}
\usepackage{algorithmic}
\usepackage{array}
\usepackage[caption=false,font=normalsize,labelfont=sf,textfont=sf]{subfig}
\usepackage{textcomp}
\usepackage{stfloats}
\usepackage{url}
\usepackage{verbatim}
\usepackage{graphicx}
\usepackage{cite}
\def\BibTeX{{\rm B\kern-.05em{\sc i\kern-.025em b}\kern-.08em
    T\kern-.1667em\lower.7ex\hbox{E}\kern-.125emX}}
\usepackage{balance}
\begin{document}
\title{An autonomous robot for pruning modern, planar fruit trees}
\author{Alexander You$^1$, Nidhi Parayil$^1$, Josyula Gopala Krishna$^1$, Uddhav Bhattarai$^2$, Ranjan Sapkota$^2$, Dawood Ahmed$^2$, Matthew Whiting$^2$, Manoj Karkee$^2$, Cindy M. Grimm$^1$, and Joseph R. Davidson$^1$
\thanks{This research is supported in part by USDA-NIFA through the Agriculture and Food Research Initiative, Agricultural Engineering Program (award No. 2020-67021-31958) and the AI Research Institutes program supported by NSF and USDA-NIFA under the AI Institute: Agricultural AI for Transforming Workforce and Decision Support (AgAID) (award No. 2021-67021-35344).}
\thanks{$^1$Collaborative Robotics and Intelligent Systems Institute (CoRIS), Oregon State University, Corvallis OR 97331, USA {\tt\footnotesize \{youa,parayiln,josyulag,cindy.grimm,joseph.davidson\}\\@oregonstate.edu}}
\thanks{$^2$Center for Precision and Automated Agricultural Systems (CPAAS), Washington State University, Prosser WA 99350, USA {\tt\footnotesize\{uddhav.bhattarai,ranjan.sapkota,dawood.ahmed,\\mdwhiting,manoj.karkee\}@wsu.edu}}
}%


\markboth{Journal of Robotics and Automation Letters}{You et. al}

\maketitle

\begin{abstract}
Dormant pruning of fruit trees is an important task for maintaining tree health and ensuring high-quality fruit. Due to decreasing labor availability, pruning is a prime candidate for robotic automation. However, pruning also represents a uniquely difficult problem for robots, requiring robust systems for perception, pruning point determination, and manipulation that must operate under variable lighting conditions and in complex, highly unstructured environments. In this paper, we introduce a system for pruning sweet cherry trees (in a planar tree architecture called an upright fruiting offshoot configuration) that integrates various subsystems from our previous work on perception and manipulation. The resulting system is capable of operating completely autonomously and requires minimal control of the environment. We validate the performance of our system through field trials in a sweet cherry orchard, ultimately achieving a cutting success rate of 58\%. Though not fully robust and requiring improvements in throughput, our system is the first to operate on fruit trees and represents a useful base platform to be improved in the future.
\end{abstract}


\section{Introduction}


\IEEEPARstart{T}{he} production of high-value tree fruit crops such as fresh market apples, pears, and cherries requires a large, seasonal workforce. After harvesting, the most labor-intensive orchard activity is dormant season pruning (i.e. after leaf drop)~\cite{verbiest2021}, a critical perennial operation required to maintain tree health and produce high yields of quality fruit. Pruning rejuvenates the tree, replacing unproductive wood with new fruiting sites. However, pruning is also a repetitive, strenuous, and sometimes dangerous task involving workers standing on ladders on uneven terrain with sharp cutting tools. Furthermore, although not as time-constrained as harvesting, pruning does require skilled workers that are increasingly difficult to find given the increased uncertainty in the general availability of agricultural labor~\cite{calvin2010us}. As a result of these factors --- plus rising production costs --- the tree fruit industry is highly motivated to transition to robotic pruning.

Unfortunately, outdoor agricultural environments are some of the most difficult places for robots to operate in. Common factors that negatively affect performance include variable lighting conditions, unstructured environments, and suboptimal terrain that can affect mobility. For pruning in particular, the main challenges lie in perception and manipulation. Pruning systems have, in the past, required global models of the plant in order to reconstruct a 3D model for determining pruning points, a task made difficult by how thin branches are as well as the inherent complexity of plant structures. Once the system determines the pruning points, it must maneuver a pruning implement to the desired pruning point, an operation which demands high precision, collision avoidance, and error robustness. These challenges have prevented pruning robots from achieving widespread deployment -- there are no commercially available robotic pruning systems for trees.

\begin{figure}[!t]
\centering
\includegraphics[width=\columnwidth]{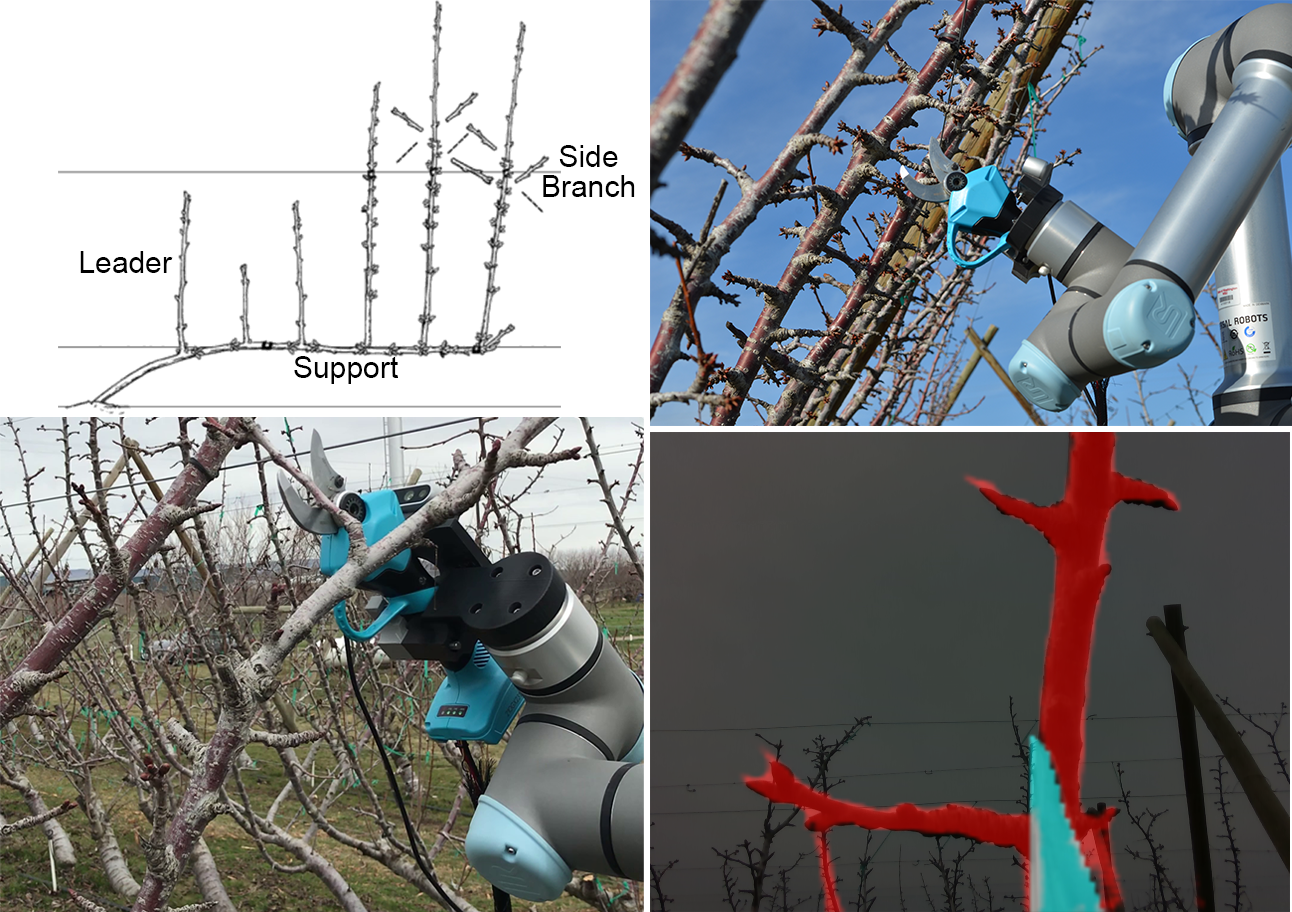}
\caption{\emph{(Top left)} A diagram of a UFO cherry tree structure, including the side branches to be pruned from the vertical leaders. \emph{(Top right and bottom left)} Our pruning robot uses a manipulator mounted on a mobile base with an eye-in-hand RGB-D sensor and electric bypass shears. \emph{(Bottom right)} Segmented image taken during the approach to the pruning point.}
\label{fig:graphicalabstract}
\end{figure}

In this paper, we present an integrated system (Figure~\ref{fig:graphicalabstract}) that advances the state of the art in sensing capabilities for pruning. Our system, once driven in front of a target tree, is capable of autonomously detecting pruning targets, moving towards them, and then executing precision cuts, representing our most complete end-to-end system to date and the first such system to be tested on fruit trees. To validate our concept, we chose an orchard system with simple pruning rules and then evaluated the robot in realistic field conditions. The system is capable of operating completely autonomously, though for our experiments we incorporate human intervention at a limited number of points for safety-critical operations. We demonstrate an overall success rate of 58\%, successfully cutting 22 out of 38 detected branches. Though the throughput is currently slow, with an average cutting time of 35.1s per cut, our system represents a solid foundation from which we can continue to improve our subsystems to achieve fast, accurate, and completely autonomous pruning.  

The paper is laid out as follows. First, we discuss prior work on pruning in Section~\ref{sec:relatedwork}. We describe the environment, orchard system, and horticultural goals of pruning in Section~\ref{sec:problemstatement}. We then provide an overview of our pruning procedure and describe in detail the hardware (Section~\ref{sec:systemoverview}) and technical components of our system (Section~\ref{sec:pruningprocedure}). Finally, we discuss the physical field trial that we conducted with the pruning system (Section~\ref{sec:systemeval}), followed by an analysis of its performance (Section~\ref{sec:results}).

\section{Related Work}
\label{sec:relatedwork}

A recent review article by Bac et al.~\cite{bac2014harvesting} discusses 50 different robotic harvesting systems developed up to 2014. Compared to harvesting, the prior work on robotic pruning remains relatively sparse, though recently there has been an increase in interest, covering aspects such as tree modeling for pruning point detection~\cite{ma2021automatic}, pruning manipulator design~\cite{zahid2020development,zhang2022design}, path planning~\cite{zahid2020collision}, and manipulator control~\cite{yandun2021reaching}. Some examples of research prototype end-to-end systems for fruit-related pruning include Botterill et al.~\cite{botterill2017robot}, Vision Robotics~\cite{VisRob}, and the Bumblebee system~\cite{silwal2021bumblebee}, all of which focus on grapevines. There has also been work on automated pruning for landscapes/gardens, such as the Trimbot2020 system~\cite{strisciuglio2018trimbot2020} that performs rose pruning and bush trimming. 

Focusing on the grapevine pruning robots, our system differs from these works in several key aspects. First, these systems focus heavily on reproducing 3D structures via stereo cameras in controlled lighting conditions to determine pruning points. In contrast, we focus on a tree architecture with relatively simple pruning rules (i.e. trimming lateral branches from the main branches). As such, we do not require 3D reconstructions of the entire tree to determine pruning points, allowing us to forgo depth-based data nearly entirely. Also, all of the previous systems featured sophisticated methods for controlling the lighting of the environment: Vision Robotics and Botterill et al. utilize a trailer towed by a tractor that completely encompasses the plant of interest, while the Bumblebee system utilizes a sophisticated lighting system to control the exposure of the images used to reconstruct the point clouds~\cite{silwal2021robust}. Similar trailer systems have been developed for fruit trees for tasks such as fruit yield measurement~\cite{gongal2016apple}. However, it is clear that using such setups significantly increases the complexity and/or unwieldiness of the system. In our work, we perform pruning in an unmodified outdoor environment by taking advantage of optical flow from camera movement to segment out important parts of the scene. This allows us to use a single RGB-D camera with no modifications to the environment.

\section{Problem Statement: Pruning rules}
\label{sec:problemstatement}


In this work, our focus is on dormant pruning of sweet cherry trees trained in an upright fruiting offshoot (UFO)~\cite{whiting2018precision} configuration (Figure~\ref{fig:graphicalabstract}). UFO trees are characterized by the presence of multiple long \textit{leader} branches that grow vertically from a horizontal support branch. The trees in our test orchard are trained in a V-shaped trellis configuration where each wall is tilted approximately 40 degrees. UFO cherry trees typically produce the most fruit on new spurs and near the base of 1-year-old shoots, and so the recommended \textit{pruning rule} for UFO trees is simply to cut off sufficiently long side branches extending from the leaders~\cite{long2015cherry}. In our previous work~\cite{you2022semantics}, we developed a skeletonization algorithm for identifying the entire structure of the tree from a global point cloud. This subsequently allowed us to extract the locations of the side branches and target potential pruning points. However, since the pruning rules are so simple for this particular orchard system, here we make the assumption that the side branches can be identified from a close-up local view of the scene without the need for sophisticated 3D modeling. We utilize this assumption about locality to run a local scanning routine that attempts to detect and cut as many lateral side branches as possible. 

\section{System Overview}
\label{sec:systemoverview}
\begin{figure}[!t]
\centering
\includegraphics[width=0.9\columnwidth]{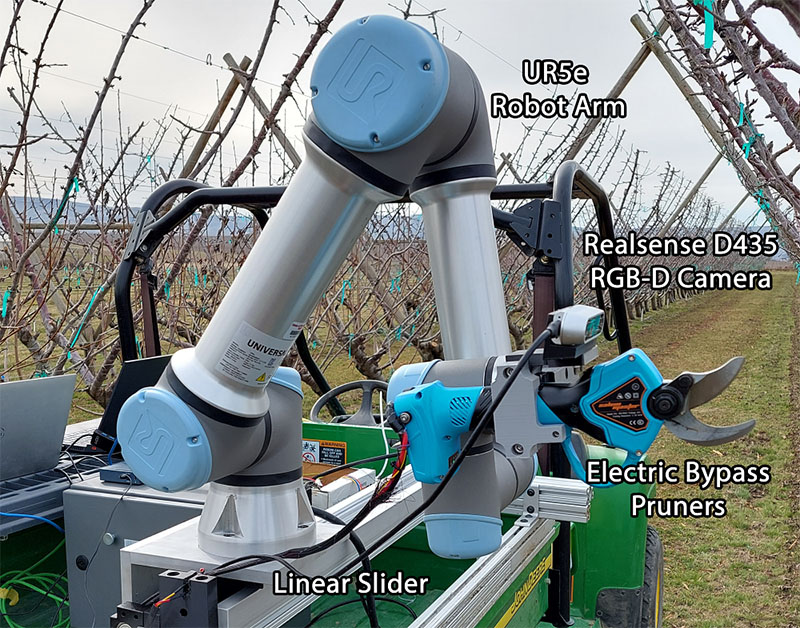}
\caption{Our pruning setup, consisting of a Universal Robots UR5e robot mounted on a linear axis. The end effector consists of a set of electric bypass pruners along with a RealSense D435 RGBD camera.}
\label{fig:hardware}
\end{figure}

Our hardware setup is shown in Figure~\ref{fig:hardware}. The robot consists of a 6 degree-of-freedom (DoF) Universal Robots (Odense, Denmark) UR5e manipulator mounted on an actuated prismatic axis (for a total of 7 DoF). The prismatic axis has a travel range of 1m and consists of a heavy duty linear slide actuated by a Nema 23 stepper motor with a lead screw transmission. Control of the prismatic axis is via a microcontroller and serial communication. For field trials, the robot is installed on the back of a utility vehicle and powered with a portable generator.

The eye-in-hand pruning end-effector integrates battery-operated electric bypass pruners, controllable via serial communication, with an Intel (Santa Clara, CA, USA) RealSense D435 RGB-D camera. The camera is located above the cutters and pitched downwards at an angle of 10 degrees so that the top blade is visible when the pruners are open. The shears are rated for cutting branches up to 3.2 cm in diameter (https://salemmaster.com/).

We control the robot using the Universal Robot's ROS driver running on Ubuntu 18.04. The computer vision algorithms (\ref{sec:maskrcnn}) were executed on a separate Dell XPS-15 laptop equipped with a NVIDIA GeForce GTX 1050 Ti graphics card running Windows 10. These algorithms represent the main bottleneck for system execution speed. Vision-based controller commands were computed on the Windows computer and sent via serial communication to the ROS computer, which in turn relayed the commands to the robot arm.


\section{Pruning procedure}
\label{sec:pruningprocedure}

\begin{figure*}[!t]
\centering
\includegraphics[width=\textwidth]{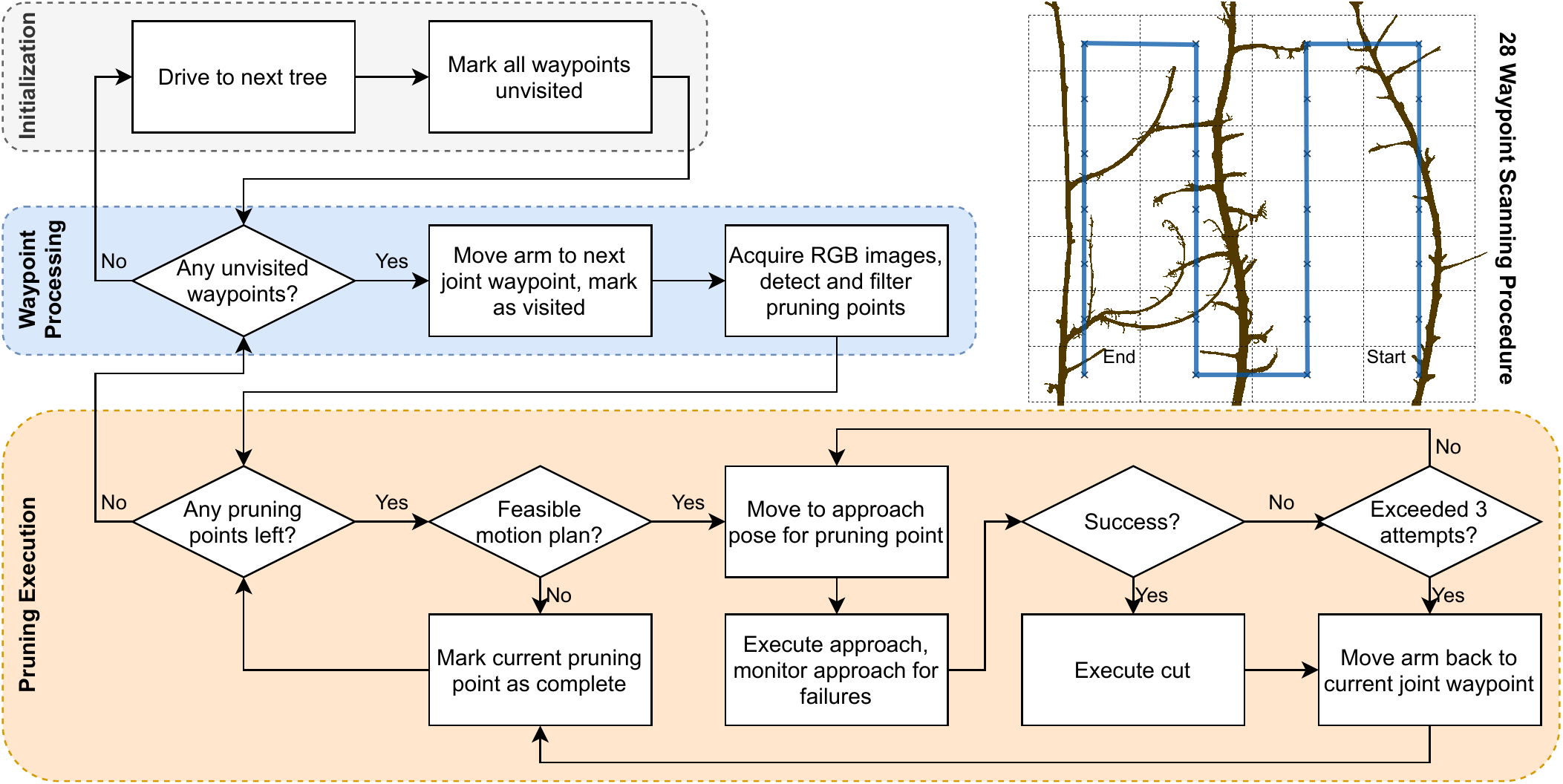}
\caption{Execution flow chart of the pruning system. At each vehicle location, we first initialize the system to its starting state. We then alternate between two steps: moving to the next waypoint and detecting pruning points, followed by operating on each of the detected pruning points. \textit{(Top right)} An illustration of the 28 waypoint scanning procedure used to locate pruning points.}
\label{fig:pruningprocedure}
\end{figure*}

In this section, we discuss the operation of our pruning system and the various essential subcomponents/algorithms. Our approach is motivated by two key factors. First, trees are wider than the field of view for a single camera positioned on the vehicle, requiring scanning. Second, we would like to reduce kinematic path planning challenges. This led us to a ``pan and scan'' through a set of fixed waypoints (upper right of Figure~\ref{fig:pruningprocedure}) to identify potential pruning points. For each potential pruning point we start at the waypoint and use a hybrid controller to execute the actual cut.

Figure~\ref{fig:pruningprocedure} shows the execution flow chart for pruning. At each pruning location (i.e. the vehicle is parked in front of a tree), the robot moves to a set of waypoints, performing a scanning motion across the trellis. At each waypoint, the robot acquires a set of two RGB images that are used to detect potential pruning points in the environment. Once pruning points are identified and converted to 3D position estimates, the robot plans to an approach pose in front of the target. Once there, the robot executes a hybrid controller that uses feedback from the RGB camera and the robot's force-torque sensor to guide the cutters to enclose the target branch, at which point we execute the cut. We include human intervention at this step to prevent unintended cuts to the tree or trellis. 

In the following sections, we describe the details for each of the system components. We assume that the vehicle has already been driven to a suitable location for pruning.

\subsection{Waypoint movement}

Our scanning routine visits a set of 28 predetermined waypoints in a zigzag shape (Figure~\ref{fig:pruningprocedure}). Each waypoint consists of one of four positions for the prismatic axis (0, 0.2, 0.4, 0.6m) and one of seven pre-configured manipulator poses. These poses are spaced 10cm apart on the vertical axis of the angled trellis plane such that the tool is oriented orthogonally to the trellis. We use the zigzag shape to avoid large joint movements between waypoints that could lead to collisions with the trellis wall. Planning between joint waypoints is done using the RRT-Connect algorithm~\cite{kuffner2000rrt}, implemented via the Open Motion Planning Library~\cite{sucan2012open} in ROS.

\subsection{Pruning point estimation}
\label{sec:prunepointestimation}

\begin{figure*}[!t]
\centering
\includegraphics[width=\textwidth]{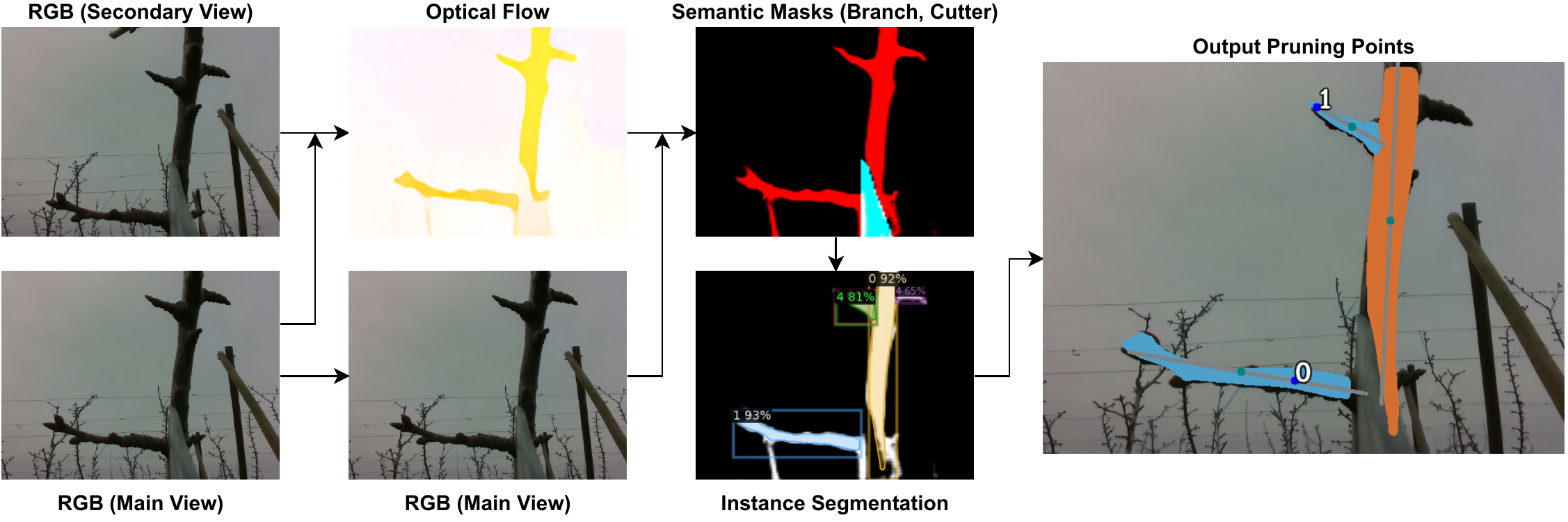}
\caption{Pipeline for the detection of pruning points. We perform a semantic segmentation of the scene (branches are red and the cutter is cyan) by utilizing optical flow data computed from 2 RGB images. We feed the branch mask into a Mask R-CNN network to perform instance segmentation of different branch components and process the resulting instances to obtain pixel estimates for pruning points. Note that in the instance above, point 1 is an example of a spur erroneously detected by the algorithm as a pruning candidate which we manually choose to skip.}
\label{fig:prunepointestimation}
\end{figure*}

Once our system has reached a waypoint, the next step is to identify the potential pruning points at that location. This is accomplished by acquiring images of the scene and feeding them through a three step process, demonstrated in Figure~\ref{fig:prunepointestimation}. In the first step, we run a semantic segmentation algorithm that identifies parts of the image that correspond to foreground tree branches (Section~\ref{sec:branchsegmentation}). We then use the branch segmentation and feed it into an instance segmentation network tasked with identifying leader branches and prunable side branches (Section~\ref{sec:maskrcnn}). We then execute an intersection detection algorithm which attempts to find intersections between candidate side branches and leaders, yielding pixel estimates for pruning points which can be converted into 3D position estimates (Section~\ref{sec:posestimates}). Notably, our pruning procedure relies on the assumption that a local view of the scene is sufficient for determining valid pruning points.  

\subsubsection{Branch segmentation}
\label{sec:branchsegmentation}

The first step in the pruning point detection pipeline is to perform semantic segmentation on the image in the current view. Our goal is to segment out two important features in the image: branches in the foreground of the image as well as the top blade of the cutter. Orchard environments can be very noisy, and often a single RGB image of a scene will not suffice to robustly determine which parts of the image are in the foreground versus the background. 

We choose to use the methodology described in our previous work~\cite{you2022optical} in which we train the pix2pix~\cite{pix2pix2017} generative adversarial network (GAN) to perform the segmentation. In lieu of using depth data to filter out the background, as is often done in other agricultural work using RGB-D cameras, we choose to compute the dense optical flow (i.e. pixel movement speed) using the SegFlow2 network~\cite{Cheng2017} between an image obtained from a second position offset 1.5cm in the y-axis of the tool frame. Once the optical flow is obtained, we stack it with the primary RGB image and feed it through the trained GAN to obtain a 2-channel output. The first channel corresponds to the branches in the scene (red), and the second channel corresponds to the cutter blades (cyan).

\begin{figure}[!t]
\centering
\includegraphics[width=0.8\columnwidth]{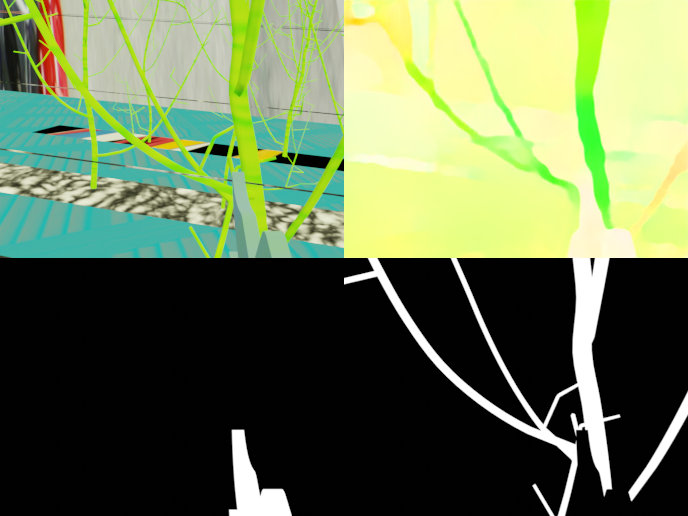}
\caption{Examples of simulated data used to train the orchard segmentation network:  RGB image (upper left), computed optical flow from a second RGB image (upper right), and the cutter and branch masks (bottom).}
\label{fig:gantraining}
\end{figure}

All of the methodologies we used were the same as before, except we included a channel in the segmentation output for the cutter. We utilized the same simulated Blender environment but included a model of the cutter in the camera view. In each randomized image, we added noise to the the pose of the cutter relative to the camera to account for imperfectly mounting the camera on the end effector. An example of the simulated data used to train the network is shown in Figure~\ref{fig:gantraining}.

\subsubsection{Instance segmentation}
\label{sec:maskrcnn}

\begin{figure}[!t]
\centering
\includegraphics[width=0.7\columnwidth]{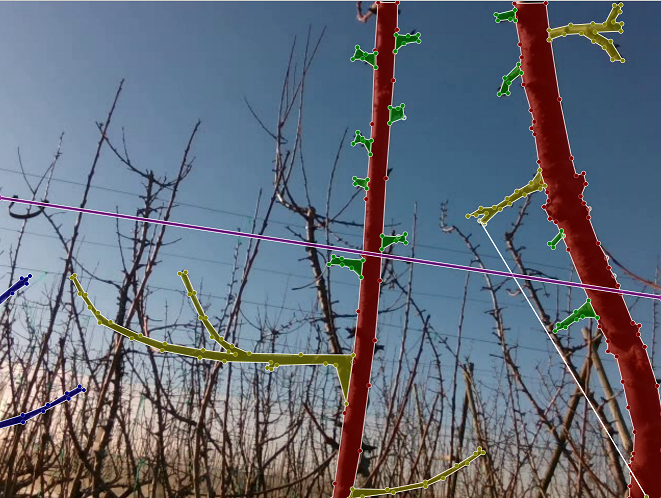}
\caption{An example of a labeled image, showing 5 labeled classes of foreground objects: leaders, side branches, small spurs, ``other" branches (usually offscreen branches that do not definitively fit in one of the former categories), and non-branch objects.}
\label{fig:labeledimage}
\end{figure}

Once we obtain the branch mask for the scene, the next step is to take the mask and identify individual branches. For this, we use the Mask R-CNN~\cite{he2017mask} instance segmentation network, provided through the Detectron2 library~\cite{wu2019detectron2}. Mask R-CNN takes in an image and outputs a set of bounding boxes and binary masks for each individually detected item in the scene. To train the Mask R-CNN, we manually labeled 371 images (also used as a comparison data set for the optical flow-based segmentation network), which were split 80\%/10\%/10\% between training/validation/testing. We labeled 5 classes of foreground objects: leaders, side branches, spurs, an ``other" branch category used for tree branches that either extended off the side of the image or did not belong in any of the aforementioned categories, and non-branch objects (primarily wires, wooden posts, and ribbons). An example of a labeled image is shown in Figure~\ref{fig:labeledimage}. For the rest of the paper, we only use the side branch and leader detection classes. 

To enable generalization of the Mask R-CNN to multiple environments, we first process all RGB images through the segmentation network (Sec~\ref{sec:branchsegmentation}) to produce a single channel mask corresponding to the branches. The Mask R-CNN network is then trained using the segmentation masks as the input. Using the masks as the network input ensures that the network focuses primarily on the shapes of the branches rather than overfitting to the specific color profile of the training data.



\subsubsection{Producing pixel and cut position estimates}
\label{sec:posestimates}

\begin{figure}[!t]
\centering
\includegraphics[width=\columnwidth]{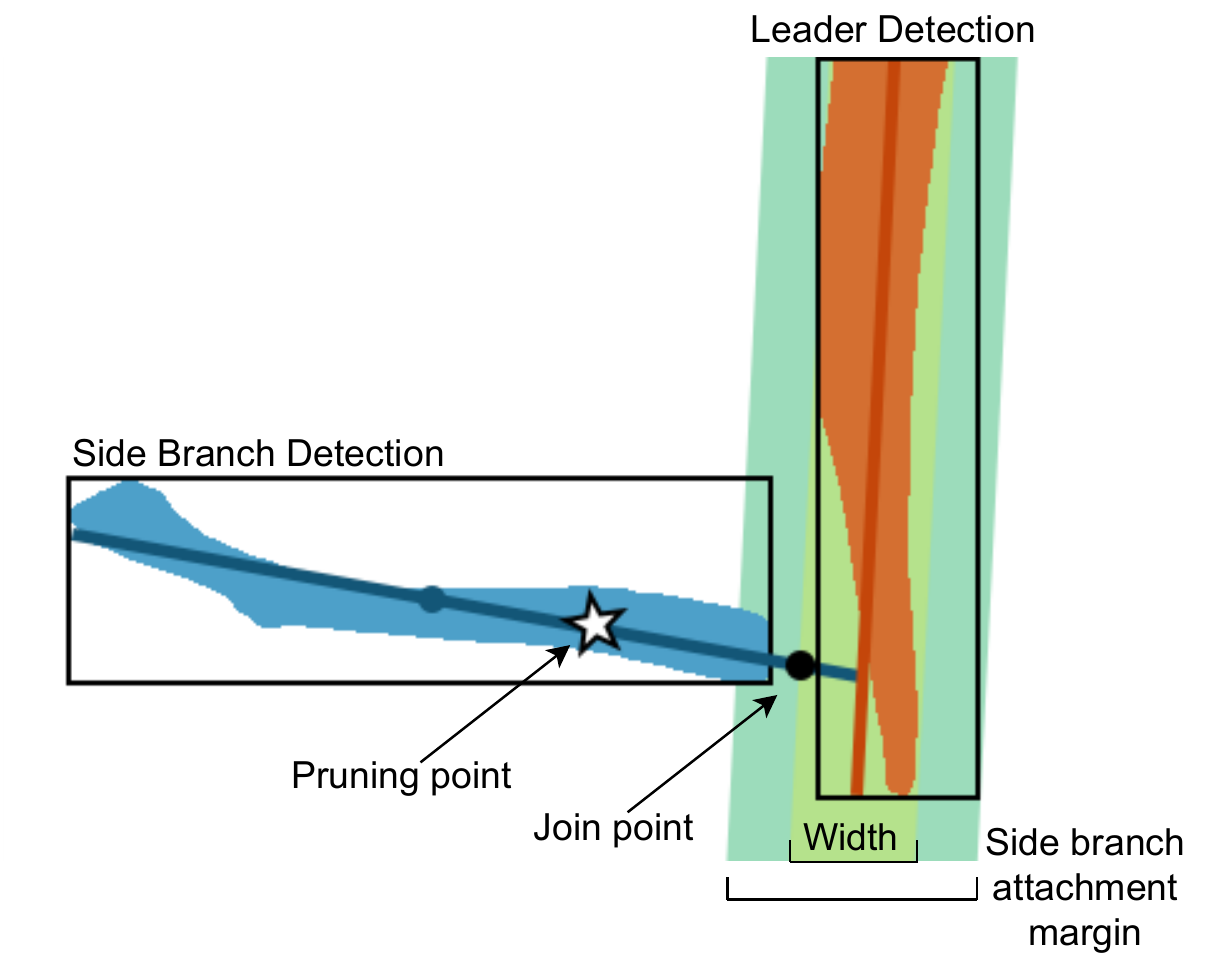}
\caption{The process of matching a detected side branch with a leader. Each detection is represented as a line segment. If a side branch intersects the leader line segment and the mask is sufficiently close to the leader, the pruning point is computed a fixed offset away from the estimated join point between the side branch and leader.}
\label{fig:branchintersection}
\end{figure}

Once the side branches and leader branches are detected, the next step is to match each branch detection to a leader (if possible) and determine the location of intersection. The method is illustrated in Figure~\ref{fig:branchintersection}. First, all of the detected masks are decomposed using PCA to yield a primary orientation and center. Each leader branch is assigned a width $w$ equal to the average number of pixels in the horizontal direction along the mask. For each side branch-leader pair, we check to see if their point of intersection $p^*$ lies inside of the leader and if any part of the side branch mask is sufficiently close to the leader boundary (at width $w$). If so, letting $\overrightarrow{b}$ represent the unit vector for the side branch (facing away from the trunk), we compute the pixel pruning point as $p^* + \left(\frac{w}{2} + m\right)\overrightarrow{b}$, where $m=90$ is a manually specified pixel offset from the \textit{join point} between the leader and the side branch. We repeat this process for every side branch-leader pair.

To convert the pixel estimate into a 3D location, we make an assumption that the pixel is located on a plane 30cm away from the camera's optical frame in the z-direction. We then use the intrinsics of the camera to deproject the point onto the plane, yielding a 3D estimate of the point in the camera's optical frame. We then use our kinematic model of the arm with the camera mounting to convert the camera frame point into a world frame estimate.

Note that the estimate of the pruning point in the world is not particularly precise, since the 30cm planar estimates for the pruning points will not always be true and the 90 pixel offset is chosen arbitrarily. Section~\ref{sec:approach} discusses the controller we use during the approach phase to correct for this estimate error.

\subsection{Moving to the approach position}

Once we convert the pixel estimates for the pruning points into a 3D target, the next step is to determine the approach pose for the robot arm. To do so, we maintain the orientation of the cutter associated with the current joint waypoint and compute the position for the end-effector such that the pruning point's position estimate would coincide with a position of $[0, 0, 0.30]$ in the end effector's frame. We then use the RRT-Connect motion planner~\cite{kuffner2000rrt} to move the robot to this pose.

\subsection{Closed-loop approach (waypoint to cut point)}
\label{sec:approach}

Once the robot moves to the approach position, the final step in the pruning process is to move the cutters towards the target branch and execute the cut. At this point, we make an assumption that the initial estimate of the pruning target is good enough so that, when the arm is moved to the approach pose, the leader and side branch are both visible. We then execute a hybrid controller that uses visual and force feedback to guide the cutter to the branch.

The details of the hybrid controller are explained in our previous work~\cite{you2021precision}, with minor changes to parameters. The visual controller is a deep neural network that takes in a segmented version of the scene and outputs a control action $(v_x, v_y) \in [-1, 1]\times[-1, 1]$. Given a forward velocity of $s=0.03$, the end effector is commanded to move at a Cartesian velocity of $[sv_x, sv_y, s]$ in the end effector's frame. We chose to run the visual controller at 1 frame per second in order to obtain reasonable optical flow estimates in between successive frames before updating the control velocity. This controller was trained in a simulated PyBullet~\cite{coumans2016pybullet} environment by formulating a pruning episode as a Markov Decision Process and training it using the Proximal Policy Optimization (PPO) algorithm~\cite{schulman2017proximal}. By utilizing the segmentation network as a means of normalizing synthetic and real images, the controller is able to transfer to a real robot with no adjustments necessary. Once a force exceeding 1.5N is detected by the force-torque sensor on the robot's wrist, we switch over to an admittance controller that regulates the control velocities to guide the branch inside the cutters while minimizing forces on the environment. Once the admittance controller terminates, we execute the cut using the electric cutters.

The segmentation scheme we use is the same one as described in Section~\ref{sec:branchsegmentation}, in which a mask is produced for the cutter and the branch in the scene using optical flow computed from successive RGB frames. We make two changes from our previous work for the control framework. First, for the network which outputs the control action, we choose to append the 3-channel optical flow to the 2-channel mask, resulting in a 5-channel input. We chose this setup due to the segmentation network sometimes leaving holes in the segmentation, especially with simulated data. The optical flow data was typically more reliable in showing the shape of the object in front and so provided a fallback for the controller in case the segmentation quality was poor. Second, because the cutter stays fixed in the frame of the view of the camera, we choose to replace the predicted cutter segmentation with a ground truth mask for the cutter. This ground truth was automatically computed by PyBullet for the simulated environment and was provided by a manually labeled mask for the real environment. We set the system up this way because an accurate representation of the cutter's location is essential for accurate pruning.

We also made a change to the reward function $\mathcal{R}$ used to train the RL system. Previously, an episode was set up so that upon success (i.e. the target branch entering a success region inside of the cutter mouth), the agent would receive a large reward, and upon failure (i.e. time running out or the branch entering a failure region outside of the cutters), the agent would receive a large penalty. Otherwise, the agent would receive an intermediate reward between 0 and $\delta$ (the timestep of the simulation) corresponding to the cutter's proximity to the target branch. One issue we sometimes observed was that the system would move the arm such that the camera lost sight of the desired target, which would cause the rest of the control to fail since we would not expect the system to guide itself to a target that it cannot see. We believe the source of this behavior was the the proximity-based intermediate reward; when the cutter is far away from the branch, no matter what action it chooses, the intermediate reward will be similar in magnitude. The behavior that we actually want is for the robot to immediately line up the visible blade of the cutter on top of the target branch. As long as the robot maintains the alignment with the blade and the branch, as the cutter moves forward, the branch should hit the inside of the cutter blade, leading to a success.

\begin{figure}[!t]
\centering
\includegraphics[width=0.7\columnwidth]{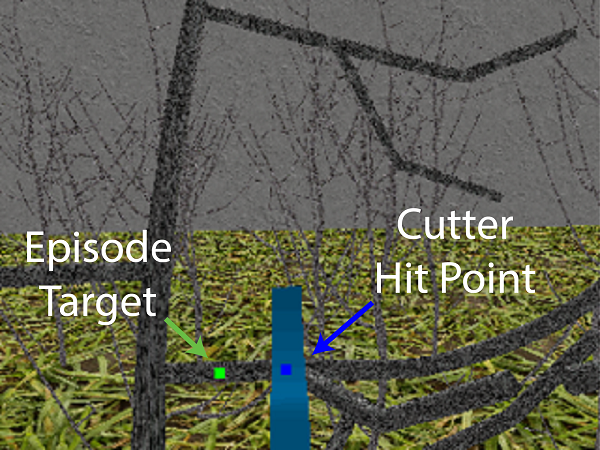}
\caption{The intermediate reward for the agent is determined by the proximity of the cutter's "hit point" and the episode target in the \textit{image} space, encouraging the agent to align the cutters with the branch quickly. Furthermore, the episode counts as a failure if the episode target leaves the image during the execution.}
\label{fig:rlreward}
\end{figure}

To encourage this behavior, we define a new intermediate reward function based on the proximity of the target and the ``hit point'' of the blade in the \textit{image} space rather than in 3D space. The basic setup is  shown in Figure~\ref{fig:rlreward}. First, we define the hit point of the blade to be at the position $[0, 0.025, 0.04]$ in the tool frame. At each step of the simulation, we project the 3D locations of the cutting target and the blade hit point into the image plane to obtain normalized image coordinates $p_{target}$ and $p_{hit}$ (i.e. locations inside the image are represented as a value in $[0,1]$). If $p_{target}$ falls outside of the image bounds, the episode instantly counts as a failure and the corresponding penalty is applied. Otherwise, the intermediate reward is based on the distance between $p_{target}$ and $p_{hit}$, as well as a manually chosen threshold $d_{thres}=0.5$ and the simulation timestep $\delta$:

\begin{equation}
    \mathcal{R}_{s_{t+1}} =\delta \max\left(0, 1 - \frac{\|p_{target} - p_{hit}\|}{d_{thres}} \right)
\end{equation}

Changing the reward function like this encourages the system to align the cutters with the branch as quickly as possible. Combined with the additional failure criterion of the target branch leaving the image, we were able to achieve more robust and consistent performance out of the approach phase. 

\section{System Evaluation}
\label{sec:systemeval}


We evaluated our system at a UFO sweet cherry orchard in Prosser, WA on an overcast day on March 17, 2022. As it was late in the dormant season, these cherry trees had already been partially pruned by workers. Despite this, we identified 10 locations in the orchard which had a suitable number of eligible branches for pruning. For a location to be eligible, there had to be at least 2 leaders located within the scanning range of the linear axis, each with at least one candidate branch for pruning. To avoid collisions, we manually pre-pruned any branches sticking out from the trellis wall.

After resetting the linear axis to the first joint waypoint (i.e. linear axis set to the zero position), a human operator drove the vehicle so that the end-effector was placed at an appropriate distance from the trellis wall (about 20-30 cm) and the first leader was centered in the view of the camera. Once the vehicle was positioned we executed our 28-waypoint scanning procedure.


Our system was designed to involve minimal human intervention and theoretically be operable without human intervention. For this trial, we chose to manually control and monitor the system during execution for any safety-critical issues. Our three manual interventions were:

\begin{itemize}
    \item Driving the vehicle: Currently, the vehicle is driven by a human operator and not autonomously. This is important because positioning the vehicle an ideal distance from the trellis wall is nontrivial due to the necessity to avoid arm-tree collisions as the vehicle is moving, as well as the difficulty involved in stopping the differential drive vehicle an ideal distance away from the wall. If the vehicle is too close, the arm may not have enough room to maneuver, while if the vehicle is too far, the arm may not be able to reach the desired target.
    \item False positive pruning point detections: When the detection algorithm identified small fruiting spurs (removing these would decrease the grower's yields) we did not execute the cuts. No corrections were made for false {\em negative} detections, i.e. if the pruning point system failed to detect an eligible branch for pruning.
    \item Approach execution monitoring: In the event that the cutters missed the target branch or were about to collide with the environment during the approach, we manually terminated the approach and rewound the robot back to the start of the approach. For each pruning target, we allowed 3 attempts to successfully cut the target before moving on to the next target.
\end{itemize}






\section{Results and discussion}
\label{sec:results}

\subsection{Detection and cutting accuracy}

In total, we detected 38 branches that were long enough to cut. Out of these 38 branches, the robot was ultimately able to cut 22 of them, representing a 58\% cutting success rate. The breakdown of the causes of failure is as follows:

\begin{itemize}
    \item Six failures were due to a motion planning failure to the approach position. If the produced motion plan had a sum of all joint displacements above a given threshold (in our case $\pi$ radians), likely indicating a poor quality motion plan, we also counted the plan as a failure.
    \item Of the remaining 32 branches, 10 of the attempts failed due to exhausting the 3 attempt limit to reach the target, representing a 69\% success rate in ultimately reaching the branch. 
    \item Of the remaining 22 branches, the average number of attempts it took to reach the target was 1.4.
\end{itemize}

Various factors accounted for the robot missing the target. In some situations, the initial estimate obtained using an assumption of a distance of 30 cm was sufficiently inaccurate that the cutting target would be barely visible or out of sight from the approach position. Other than that, most misses were caused by the end-effector passing underneath the target branch; we observed that our RL-trained controller sometimes failed to move the end-effector up at critical moments.

Regarding the accuracy of the branch detection, as previously noted, we had 38 true positive detections of pruning points. By far the biggest issue we had with the system was the presence of false positives. In total, we had 115 false positives reported. Ninety-six of these were simply spurs being detected as side branches. We attribute this to our views in the training data being further away from the trellis wall than during our trials; since the camera was closer up during the trials, the spurs appeared to be of sufficient length in the images to be classified as side branches. In practice, this issue could be addressed by using depth data to measure the length of each detection and using a length-based cutoff to filter out detected spurs. Ten of the false positives were due to horizontal trellis wires being detected as side branches, indicating a need to more robustly model the environment. The remaining 9 false positives were mainly due to false detections of non-existent leaders leading to a spurious intersection. 

We also had 27 instances of false negatives in which an intersection of a side branch and a leader was not detected. Two of them were due to a leader not being detected. For all other 25 instances though, both the leader and the side branch were properly detected, and so the lack of intersection was due to implementation issues in our intersection algorithm, which we will address for subsequent trials.


\subsection{Runtimes}

\begin{figure}[!t]
\centering
\includegraphics[width=\columnwidth]{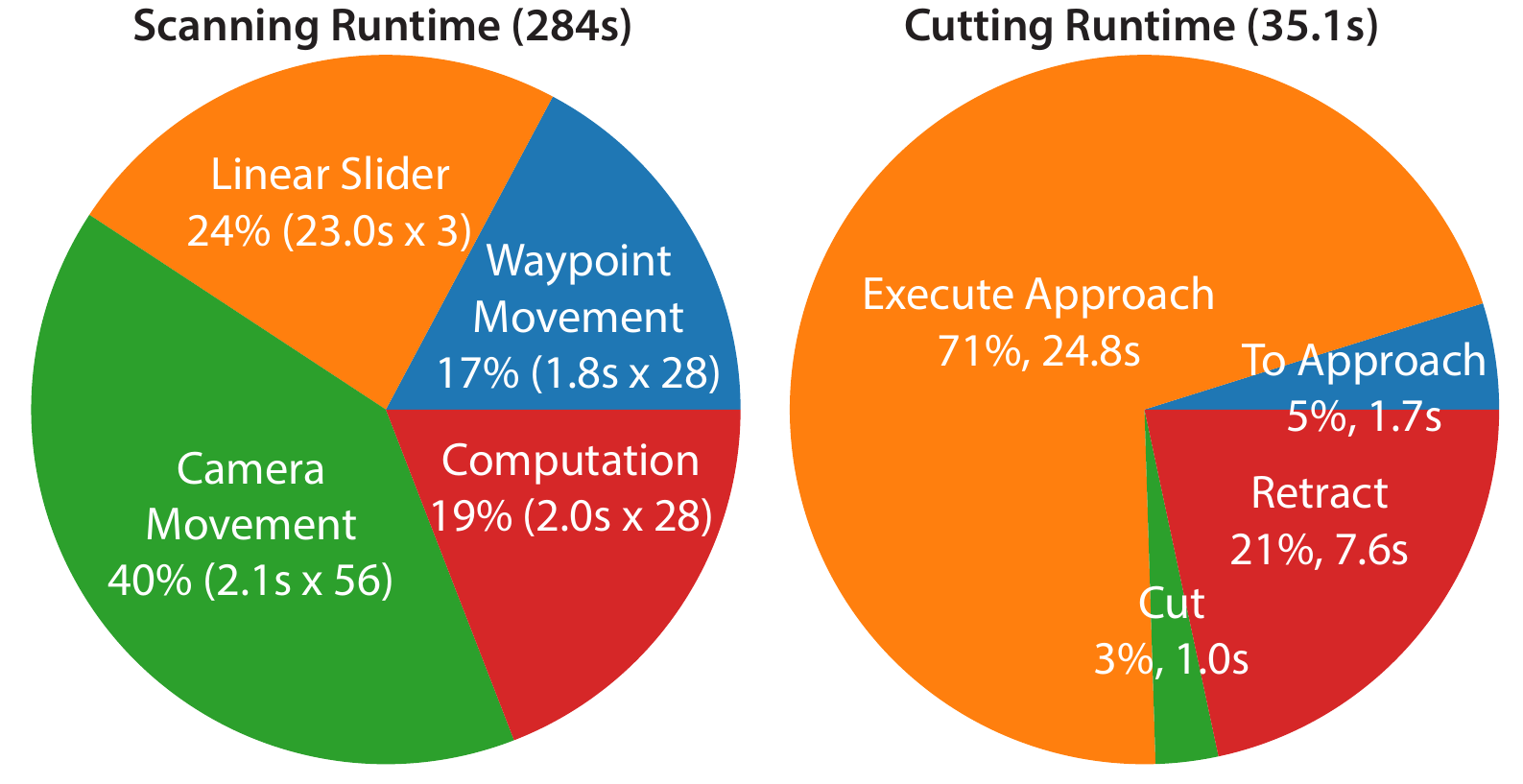}
\caption{Left: Average time spent in the different stages of the scanning procedure, not including operations on detected pruning points since the number of cuts and cutting attempts varied per tree (total 284s average). Right: Average time spent at each stage for a single cutting attempt assuming the approach succeeds on the first try (total 35.1s average).}
\label{fig:runtimes}
\end{figure}

Figure~\ref{fig:runtimes} shows the breakdown of the system's cycle time. Since the number of cuts executed is variable across experiments, we choose to analyze the runtimes of the scanning process and of a single cutting attempt separately.

Overall, an entire scan of a region spanning 0.6m horizontally and 0.6m vertically took 284s. The majority of the time (115s) was actually spent moving the camera at each waypoint (to obtain alternate views). This was because we had to move the robot arm slowly to prevent the arm from vibrating when the arm reached its goal, which we found often led to poor optical flow estimates. However, there was also some redundant movement because the robot would first move to the scan waypoint, then to the offset view, and then back to the scan waypoint (rather than continuing to the next scan waypoint). By eliminating this redundancy, we could cut down on the time spent moving the camera by half.

The other notably slow part of the scanning procedure was moving the linear axis, which took about 23s to move 20 cm, resulting in an additional 69s to the scanning procedure. The slow movement speed was chosen partially due to safety concerns, but is also limited somewhat by the screw-based design of the system. Future iterations of our work will focus on increasing the design's travel speed. The remaining sources of time are from moving the arm between waypoints (50s) and running the detection algorithms (49s), which are not notably inefficient, though the speed of FlowNet2 and Mask R-CNN could be improved by using a more modern graphics card. 

For the cutting process, assuming success on the first try, an average cut took 35.1 seconds, consisting of 1.7s to move the arm to the approach position, 24.8s to execute the approach, 1.0s to execute the cut, and 7.6s to retract the arm back to the starting position. The main constraint to the approach was the 3 cm/s limit on the forward velocity of the end-effector, which we imposed for safety and monitoring purposes. One main contributor to the approach time was waiting for the admittance controller to terminate, as in some instances the desired balance of forces never occurred, leading us to wait for the 15s timeout on that part of the controller. Increasing the velocity of the approach would also be possible. However, moving the arm faster can lead to noisy force-torque estimates that may accidentally trigger the transition to the admittance controller, which is something we occasionally saw even with the 3cm/s cutoff. Additional work is required for developing a more robust criterion for switching to, and terminating, the admittance  controller.

\section{Conclusion}

In this paper, we present an end-to-end system for autonomously pruning sweet cherry trees in a modern, planar architecture. First, we use a novel segmentation method to reliably extract the foreground branches of the scene, avoiding the need to control the lighting of the environment. Using the foreground masks, we then use a Mask R-CNN network to detect pruning points. Finally, after projecting the detected pixel locations into 3D space and moving the robot to an approach position in front of the target, we utilize a closed-loop hybrid vision/interaction controller to accurately guide the cutters to the branch, allowing us to compensate for many sources of error in the original point cloud estimate while regulating interaction with the environment. Though the accuracy and cycle time of the system can still be improved, our field trials in a realistic outdoor environment demonstrate that our pipeline is a viable one for use in a real orchard. 

In addition to improving the performance of the individual subcomponents of our system, one major system aspect we would like to investigate in the future is the design of the manipulator. We observed that the UR5e's revolute joints were not suitable for use in a compact orchard space. If the robot was too close to the trellis wall, it had a difficult time avoiding collisions with the wall, while if it was too far away, the arm would fail to reach the target points. Our goal will be to explore other manipulator designs with kinematics that are better posed for modern orchard systems.

\bibliographystyle{./IEEEtran}
\bibliography{./references.bib}







\end{document}